\documentclass[10pt,twocolumn,letterpaper]{article}

\usepackage[pagenumbers]{cvpr} %

\definecolor{cvprblue}{rgb}{0.21,0.49,0.74}
\usepackage[pagebackref,breaklinks,colorlinks,allcolors=cvprblue]{hyperref}

\usepackage{graphicx, amsmath,amssymb,booktabs,url,multirow,bm,bbm,breqn,rotating,color,colortbl,subcaption,array,gensymb}
\usepackage[dvipsnames]{xcolor}

\usepackage[pagebackref,breaklinks,colorlinks]{hyperref}
\usepackage[symbol]{footmisc}

\def\paperID{8} %
\def\confName{CVPR}
\def\confYear{2025}

\begin{document}

\title{Aerial Infrared Health Monitoring of Solar Photovoltaic Farms at Scale}

\author{
Isaac Corley$^{*}$ \quad
Conor Wallace \quad
Sourav Agrawal \quad
Burton Putrah \quad
Jonathan Lwowski\\
\centerline{\hspace{-1cm}Zeitview}\\
{\tt\small {\{firstname.lastname@zeitview.com}\}}
}

\twocolumn[{%
\renewcommand\twocolumn[1][]{#1}%
\maketitle
  \centering
  \newlength{\itemheight}{
    \centering
    \vspace{-3ex}
    \includegraphics[width=0.95\linewidth]{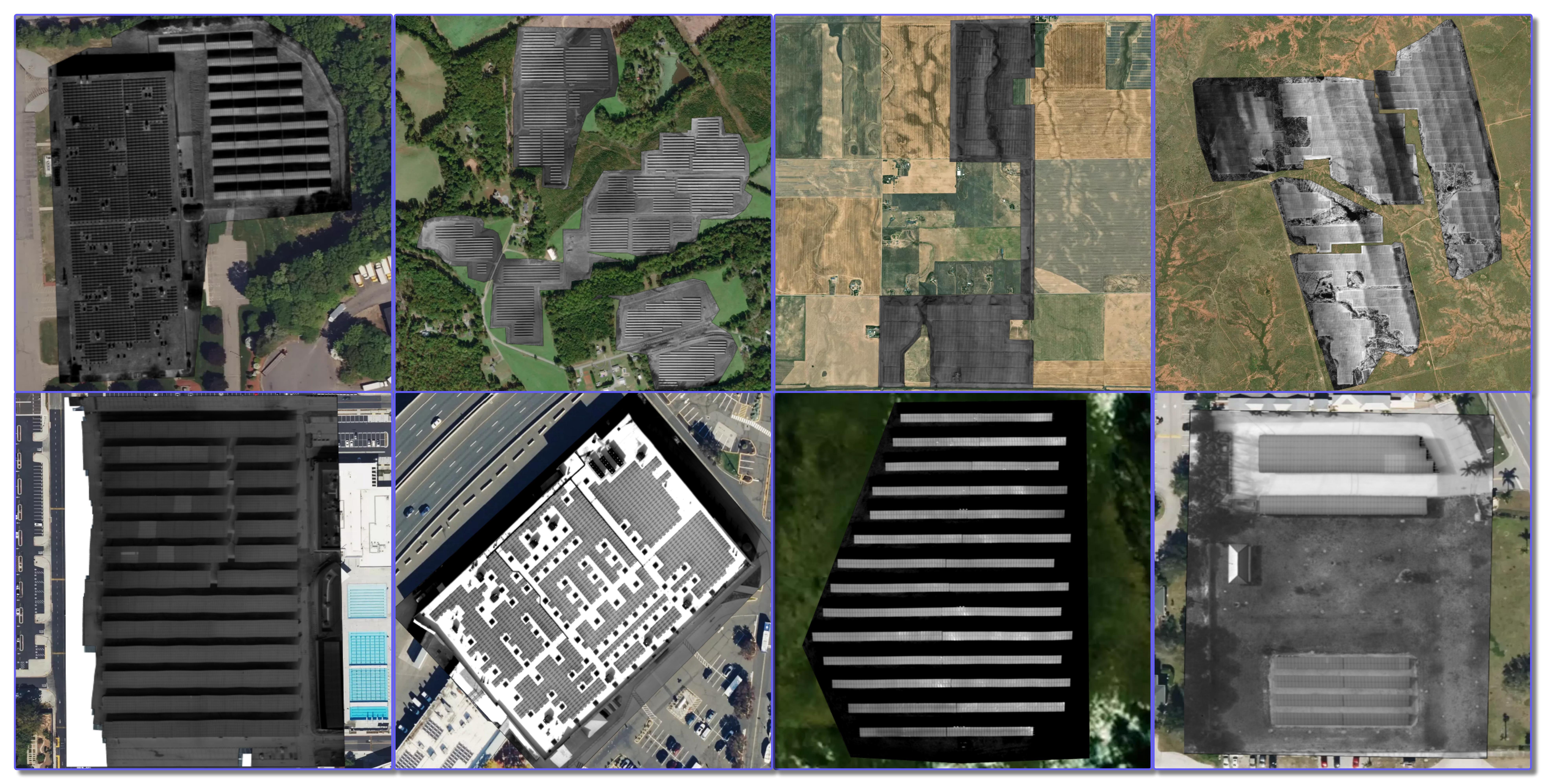}
  }
  \captionof{figure}{
  \textbf{Samples of solar sites varied by capacity in MW and panel mounting type}. \textit{Top row, left-to-right:} 1, 25, 100, and 400 total capacity in MW. \textit{Bottom row, left-to-right:} Canopy, Rooftop, Ground, and Mixed. Infrared orthomosaics are visualized in grayscale and overlaid onto RGB orthomosaics. Brighter pixels represent higher temperatures.
  }
  \vspace{3ex}
  \label{fig:sample}%
}]

\begin{abstract}

Solar photovoltaic (PV) farms represent a major source of global renewable energy generation, yet their true operational efficiency often remains unknown at scale. In this paper, we present a comprehensive, data-driven framework for large-scale airborne infrared inspection of North American solar installations. Leveraging high-resolution thermal imagery, we construct and curate a geographically diverse dataset encompassing thousands of PV sites, enabling machine learning-based detection and localization of defects that are not detectable in the visible spectrum. Our pipeline integrates advanced image processing, georeferencing, and airborne thermal infrared anomaly detection to provide rigorous estimates of performance losses. We highlight practical considerations in aerial data collection, annotation methodologies, and model deployment across a wide range of environmental and operational conditions. Our work delivers new insights into the reliability of large-scale solar assets and serves as a foundation for ongoing research on performance trends, predictive maintenance, and scalable analytics in the renewable energy sector.

\end{abstract}

\section{Introduction}
 \footnotetext[1]{Corresponding author: isaac.corley@zeitview.com}

Large-scale solar photovoltaic (PV) installations are expanding at an accelerated pace, driven by cost reductions and supportive policy measures. These multi-megawatt solar farms require frequent performance assessments to identify operational inefficiencies and mitigate safety risks—most notably, thermal anomalies that can lead to potential fires. With the passage of the Inflation Reduction Act of 2022~\cite{bistline2023emissions}, the United States is poised to further increase its solar energy capacity, intensifying the demand for scalable inspection and monitoring solutions.

Traditionally, health monitoring of these PV farms involves operators reliant on manual ground-based inspection protocols. While effective for small installations, this approach becomes infeasible as site footprints and module counts exponentially increase~\cite{wallace2023solar}. The growing number of high-capacity solar plants necessitates a paradigm shift toward automated, data-driven methods that can rapidly and accurately assess large swaths of PV modules. Our work addresses this need by introducing an aerial infrared inspection pipeline underpinned by machine learning, designed to streamline operational checks and facilitate predictive maintenance across expansive solar assets.

\vspace{-2ex}\paragraph{Motivation}\hspace{-2ex}
Global solar PV capacity has experienced a remarkable annual growth rate in recent years~\cite{mooney2009electricity}, yet critical questions remain regarding the aggregated performance of these assets and the influence of geographic or material variations over time. Understanding how PV plants degrade, lose efficiency, or develop operational defects has direct ramifications on power output, operations, and maintenance strategies.

\begin{figure}[t!]
\centering
\includegraphics[width=1.0\linewidth]{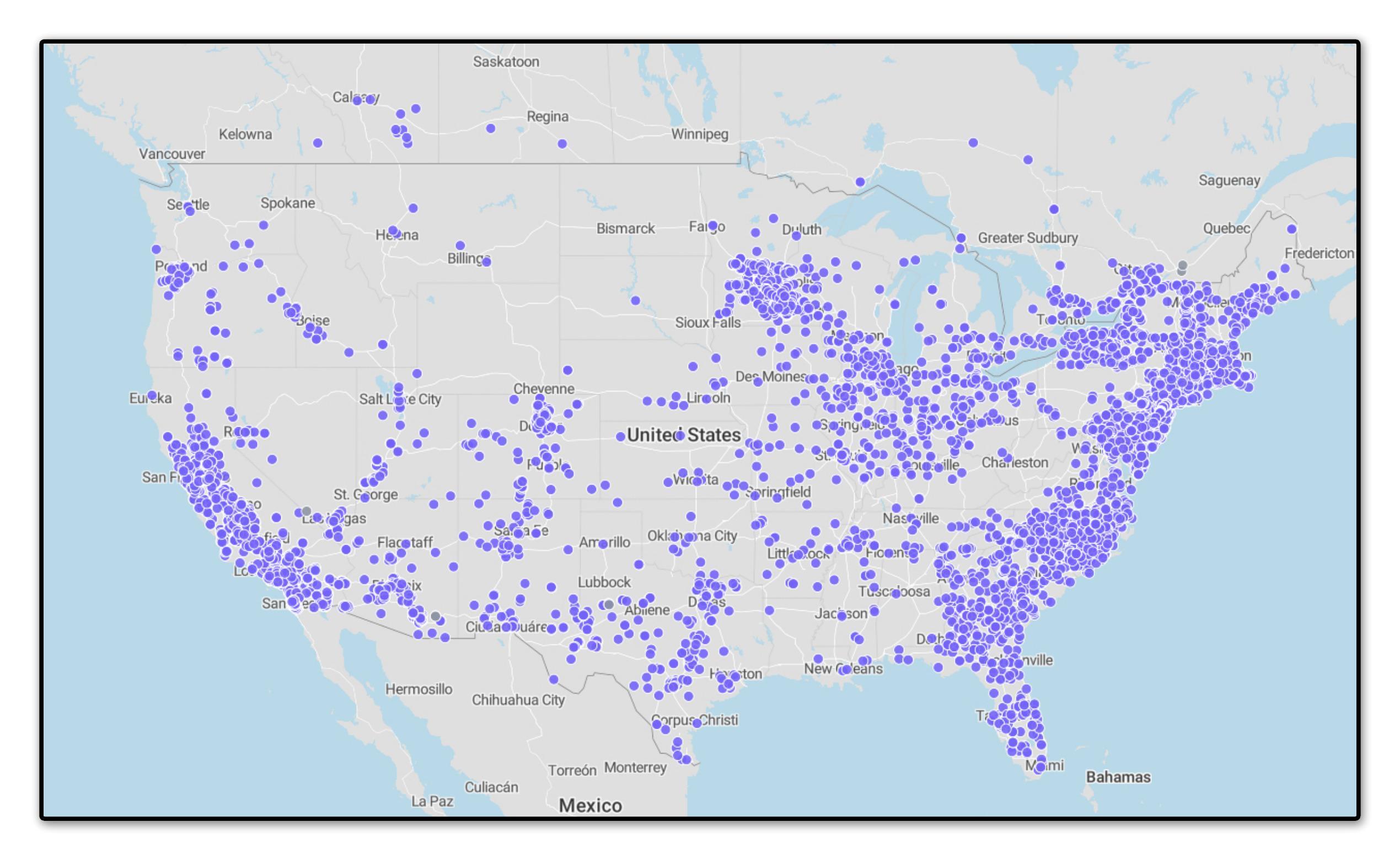}
\caption{\textbf{Geographic locations of airborne infrared inspections in our dataset.} Our dataset consists of high-resolution infrared orthomosaics of a diverse set of 6,155 solar farms across North America captured by fixed-wing aircraft.}
\label{fig:locations}
\end{figure}

Aerial inspection methods—particularly those leveraging thermal infrared imaging—have proven effective for identifying defective modules, diodes, and wiring issues. The convergence of small uncrewed aerial systems (sUAS) equipped with compact IR sensors and larger crewed aircraft outfitted with high-resolution mid-wave IR and Electro-Optical (EO) payloads has catalyzed a shift in data acquisition, transforming it into a commercially viable, industry-scale endeavor. These multi-modal data streams capture high-fidelity imagery across expansive PV installations, enabling precise, rapid evaluation of system health.

Early adopters, such as the National Renewable Energy Laboratory (NREL), have employed our airborne infrared datasets to correlate observed thermal anomalies with time-series performance data at major solar sites~\cite{perry2024relating}. This correlation underscores the potential of large-scale aerial inspection pipelines to drive both predictive maintenance and broader insights into the life-cycle performance of PV assets, providing a compelling foundation for continued innovation in machine learning-driven solar inspection methodologies~\cite{mooney2009electricity}.

\begin{figure}[t!]%
    \centering
    \begin{subfigure}[b]{0.47\linewidth}
        \includegraphics[width=\linewidth]{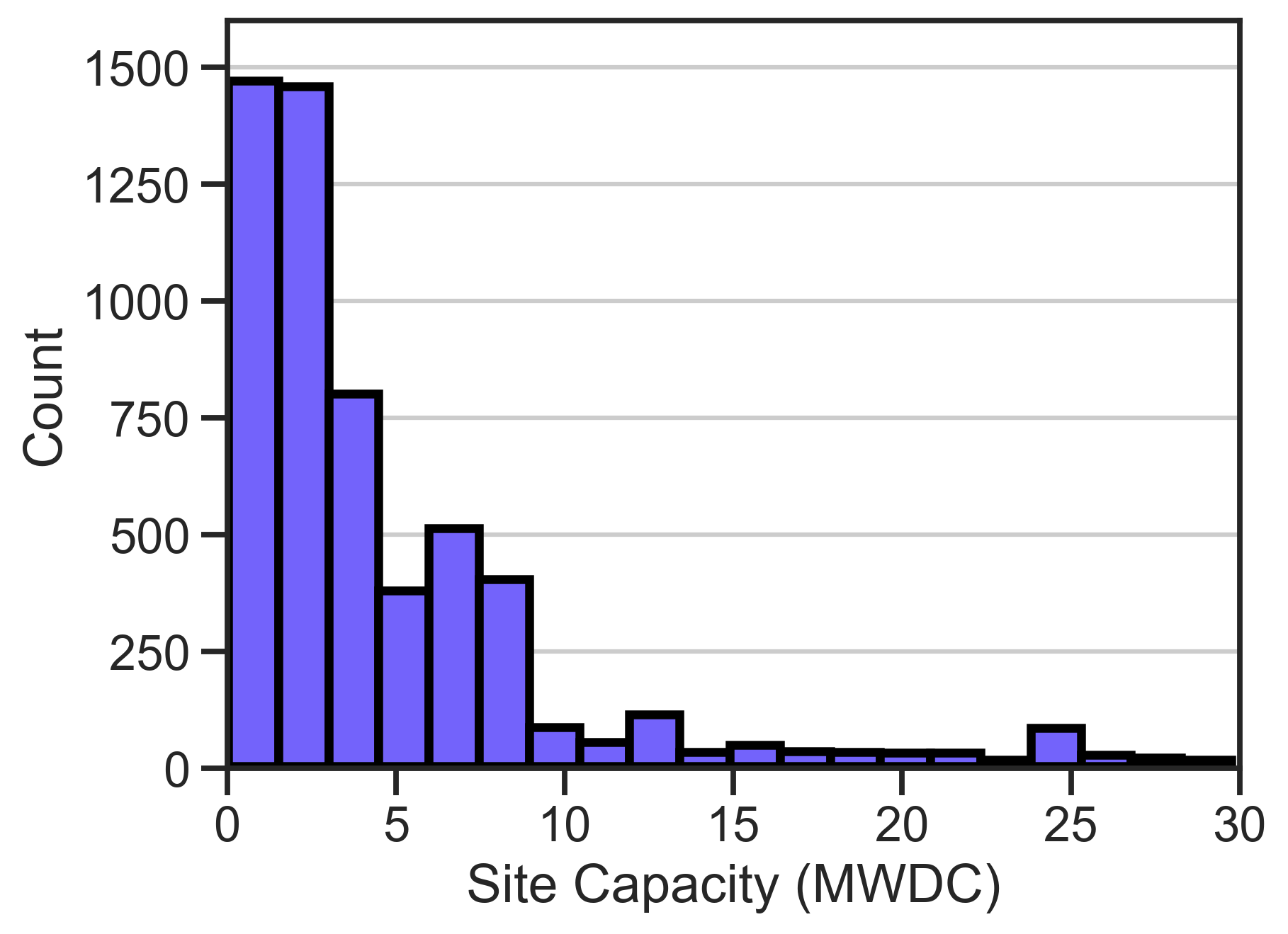}
        \caption{}
    \end{subfigure}
    \begin{subfigure}[b]{0.47\linewidth}
        \includegraphics[width=\linewidth]{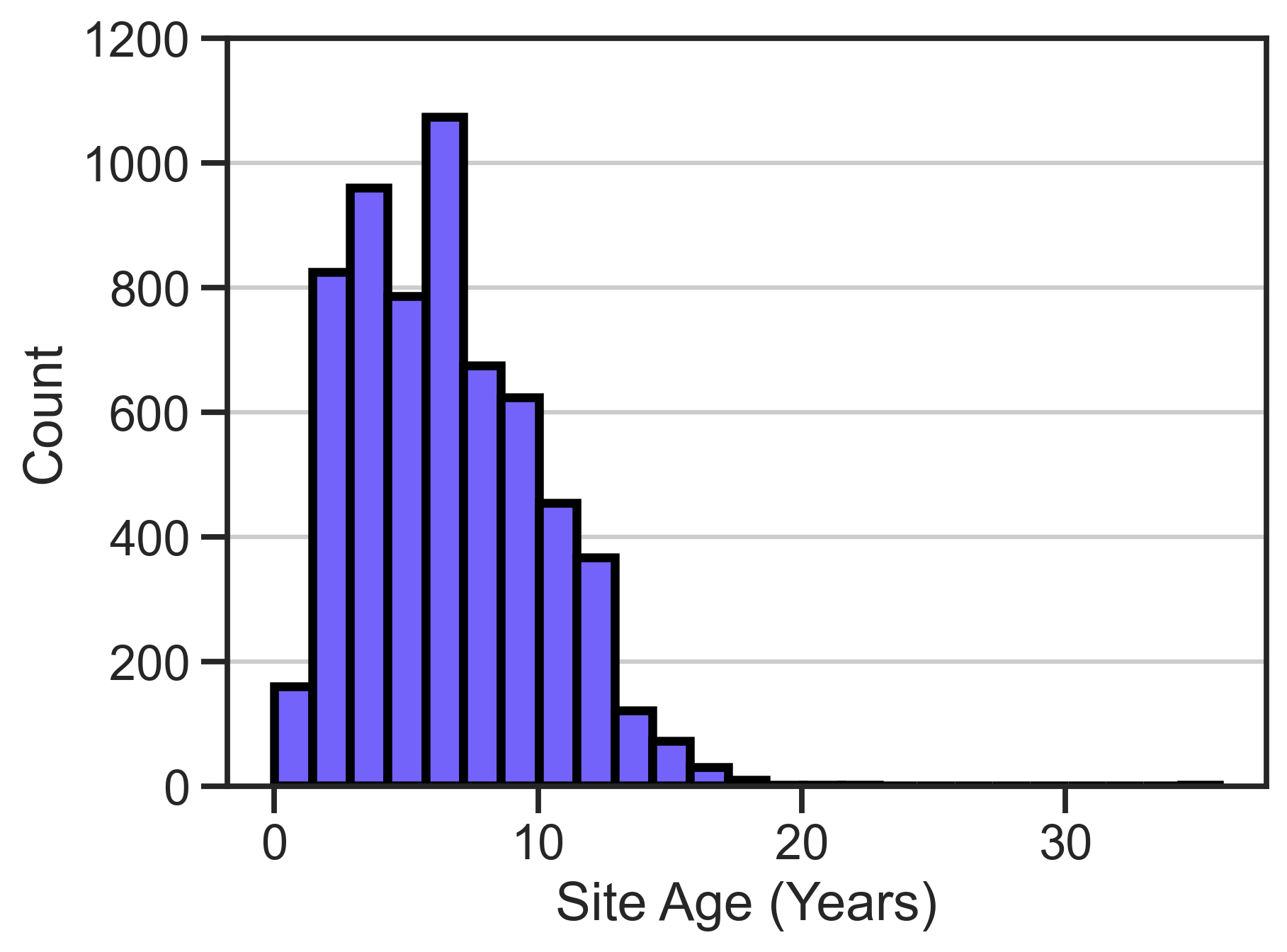}
        \caption{}
    \end{subfigure}
    \\
    \begin{subfigure}[b]{0.47\linewidth}
        \includegraphics[width=\linewidth]{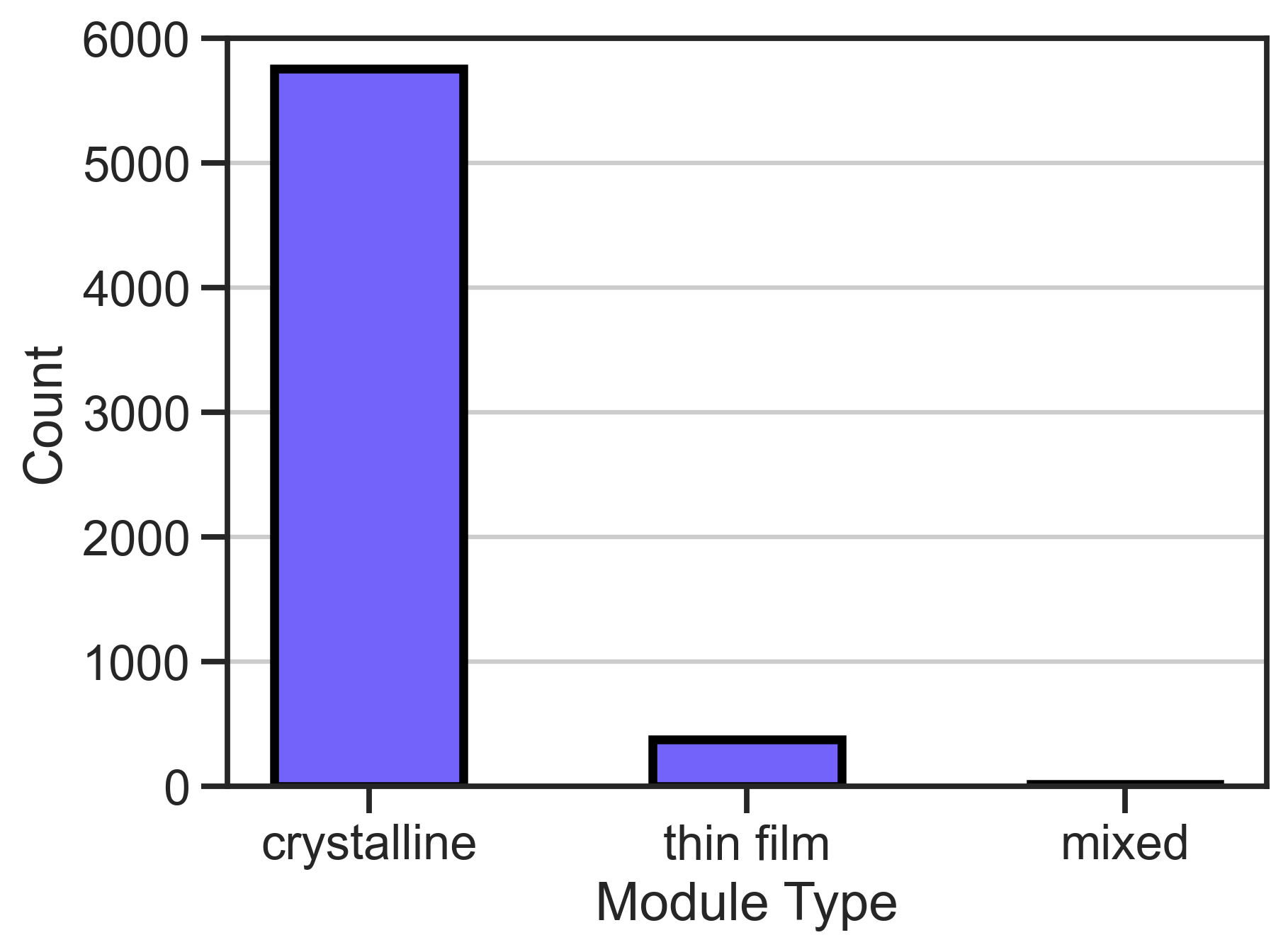}
        \caption{}
    \end{subfigure}
    \begin{subfigure}[b]{0.47\linewidth}
        \includegraphics[width=\linewidth]{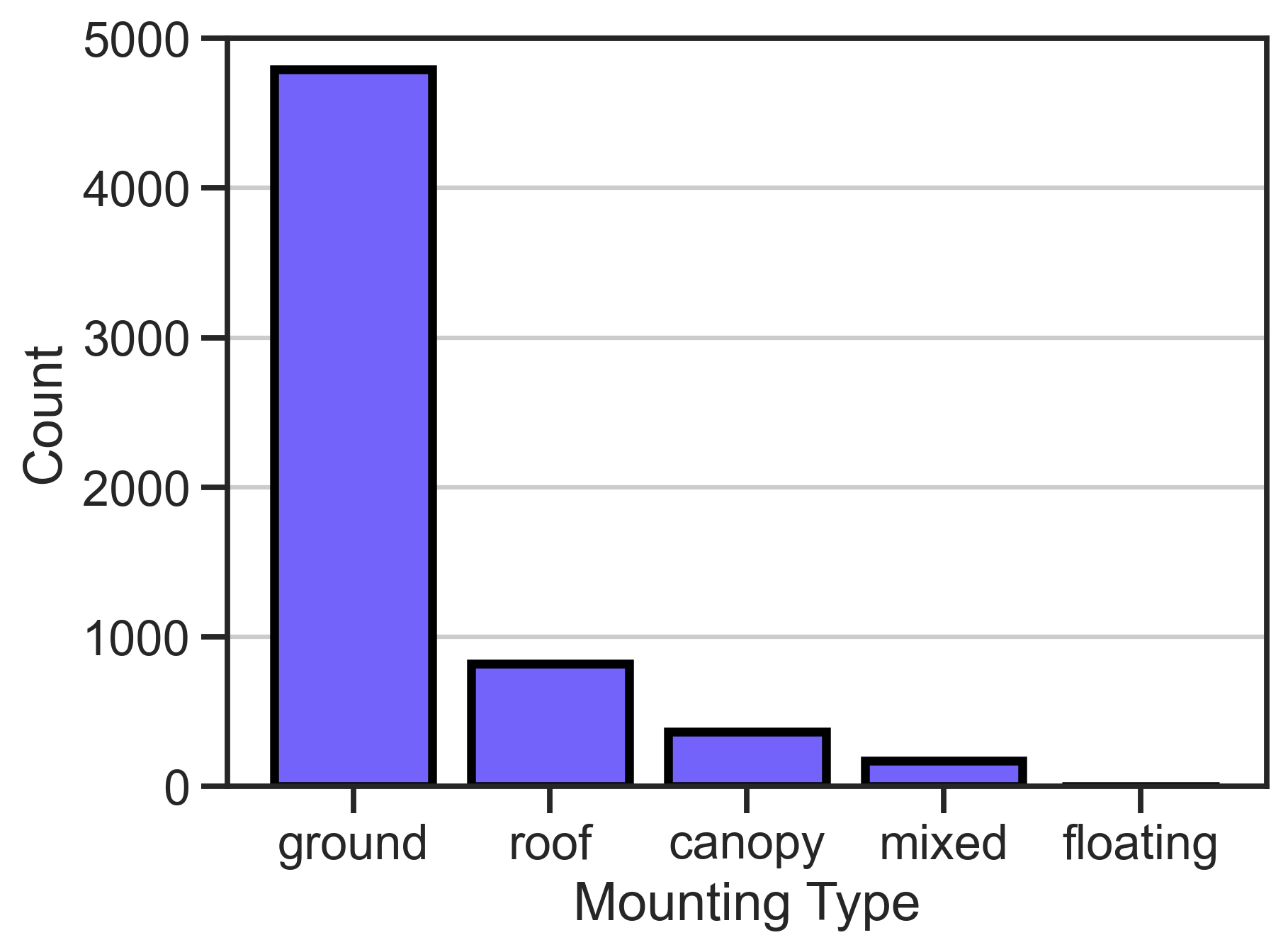}
        \caption{}
    \end{subfigure}
    \caption{\textbf{Distribution plots of dataset statistics} of (a) total site capacity in MWDC, (b) solar plant site age in years, (c) solar panel module types, and (d) solar panel mounting types.}
    \label{fig:stats}%
\end{figure}

\vspace{-2ex}\paragraph{Contributions}\hspace{-2ex}
Building on these motivations, our work advances solar asset monitoring and defect detection through the following technical innovations:

\begin{itemize}
    \item We build a high-fidelity dataset of North American solar sites ($\geq$1 MW), integrating multi-spectral, georeferenced imagery with precise temporal and environmental annotations.

    \item We introduce an automated, end-to-end deep learning pipeline for real-time anomaly detection and renewable energy loss estimation across large-scale PV plants.

    \item We propose a novel PV health rating system that combines operational efficiency, thermal anomaly analysis, and defect density into a robust, intuitive assessment tool.

    \item We perform an insightful analysis of our data to quantify potential energy and revenue losses due to identified defects, supporting informed maintenance and investment decisions.
\end{itemize}

\subsection{Related Work}
There are many prior works on mapping defects in PV plants~\cite{de2022automatic,yang2024survey, yahya2022applied}; however, none of them perform the analysis with a geographically diverse dataset or have been deployed at scale.

\begin{figure*}[t!]
\centering
\includegraphics[width=0.95\linewidth]{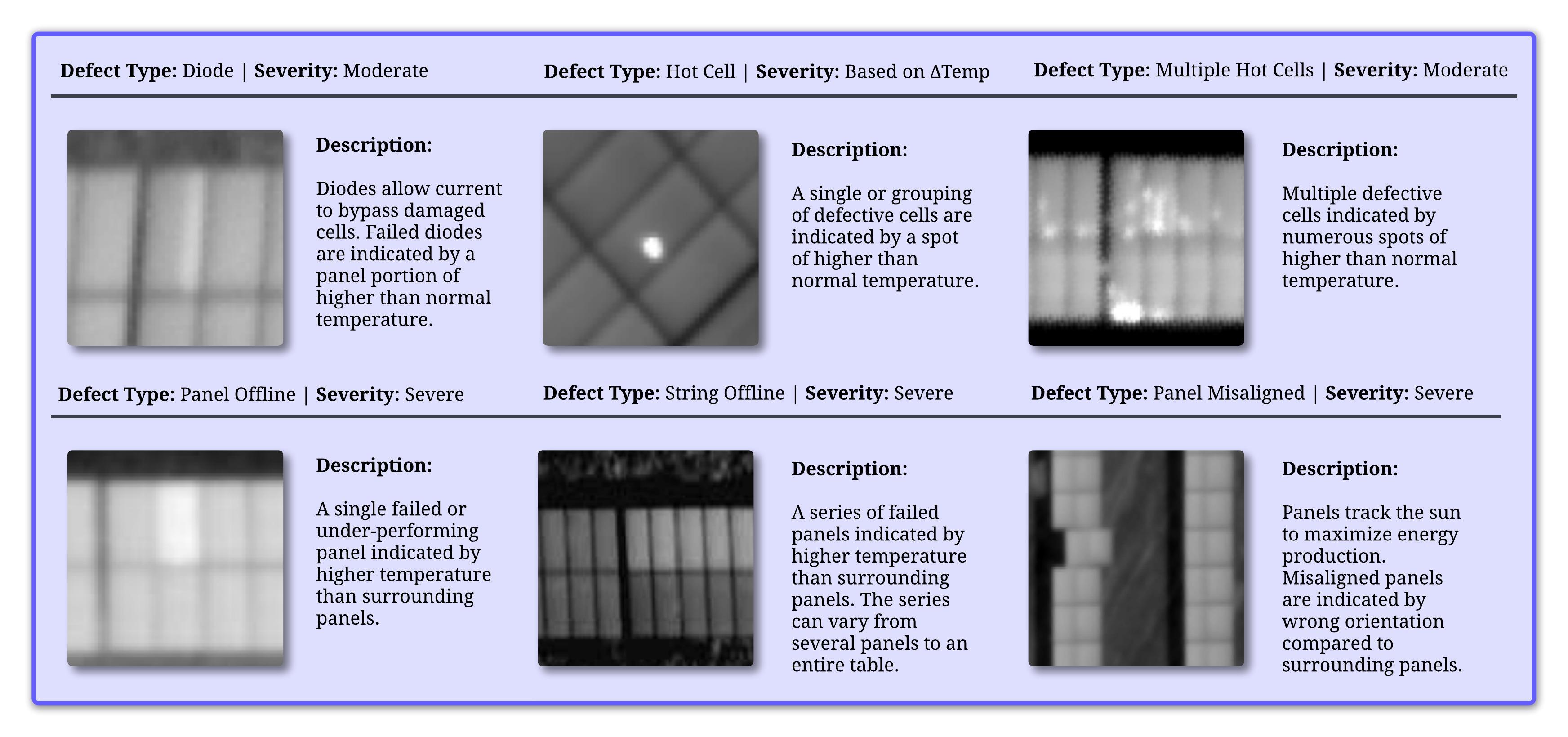}
\caption{\textbf{Solar Panel Defect Categories and their Descriptions.} For Anomaly Detection in infrared imagery, we detect 6 types of defects, including hotspots, multi-hotspots, diode bypass, single panel outage, string outage, and misaligned panels due to faulty trackers.}
\label{fig:defect-types}
\end{figure*}

\citet{millendorf2020infrared} proposed the popular benchmark dataset InfraredSolarModules of 20k $24 \times 40$  infrared crops of panels for detecting 11 defect categories; however, it's standard industry practice to have global information of nearby panel temperatures when determining if a hot area on a panel is, in fact, a defect which can't be done at the crop level. Furthermore, we experiment with InfraredSolarModules and find that \textit{a Random Forest baseline can achieve 84.0\% overall accuracy for anomaly classification} which could indicate the dataset is already too simple to deploy a model trained on this dataset at scale\footnote{Code for the InfraredSolarModules experiment is available \href{https://gist.github.com/isaaccorley/9424f71996b18e4b00b3ac54824d8e77}{here}}. \citet{wang2024pvf} follows suit by creating the PVF-10 dataset of 8k infrared crops of defective panel samples from 8 PV farms which, when deployed at scale, would be an insufficient sample size to create a model robust to geographical variances in temperature.

Other methods tend to focus on the design of the neural network architectures, and while they may show promising results with 90\%+ accuracies, they still are trained on the same datasets containing a small amount of PV plants. \citet{duranay2023fault} uses an EfficientNet-B0 image encoder combined with a Support Vector Machine classifier to achieve 93\% accuracy for detecting defects on a dataset of 20k solar panel crops. \citet{alves2021automatic} constructs a custom Convolutional Neural Network (CNN) to achieve 92.5\% accuracy on the InfraredSolarModules dataset. \citet{barraz2025cascading} utilizes a cascading ensemble of classifiers with a voting CNN model to achieve 88\% on InfraredSolarModules. \citet{gopalakrishnan2024nasnet} uses a combination of the NASNet CNN architecture with a Long-Short Term Memory (LSTM) network to obtain 85\% on InfraredSolarModules. While there is no shortage of proposed machine learning architectures for solar PV plant anomaly detection or classification, these works neglect to study this problem at scale.

To summarize, we find that the primary predictor to make airborne infrared PV farm inspection successful at scale is not the model architecture itself, but instead the construction of a geographically diverse dataset with varying climates and ground temperatures for machine learning-based inspections.

\section{Data}
\label{sec:data}

\paragraph{Data Acquisition}\hspace{-2ex}
Data acquisition was primarily conducted via crewed fixed-wing aircraft (e.g., single-engine Cessna 172, Tecnam P2006) outfitted with a multi-sensor payload. When adverse weather conditions, economic constraints, or local regulations prevented fixed-wing flights, sUAS were employed to maintain coverage. Missions were scheduled to maximize solar irradiance and minimize transit times between sites, with specific flight windows determined by local meteorological forecasts.

\begin{figure*}[t!]
\centering
\includegraphics[width=1.0\linewidth]{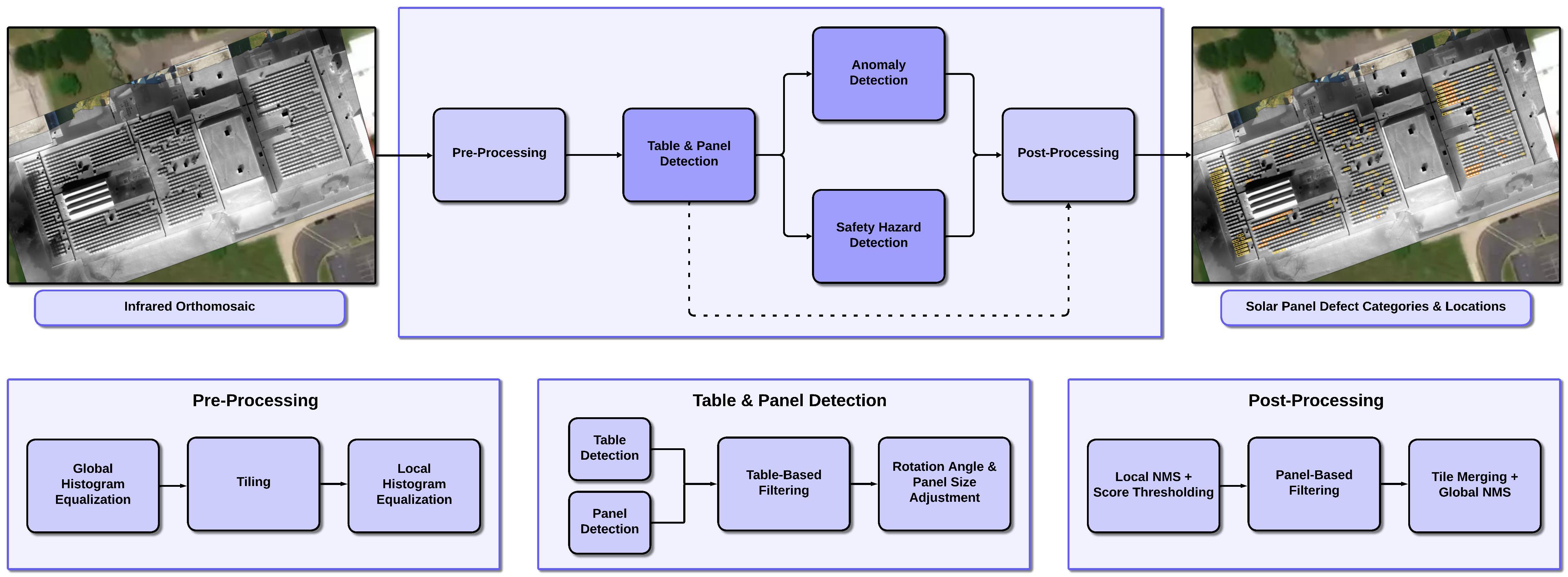}
\caption{\textbf{Architecture of our Infrared Solar Farm Inspection Pipeline.}}
\label{fig:architecture}
\end{figure*}

Each aircraft followed a systematic “lawnmower” survey pattern, flying horizontally across the PV site with overlapping passes aligned to module rows where possible. This approach maximized sensor coverage and ensured sufficient image overlap for orthomosaic generation. Infrared (IR) imagery was captured at resolutions of up to 6.5 cm ground sampling distance (GSD), while EO imagery was acquired at 2.2 cm GSD. Wider swath sensors were used on larger airframes to reduce the total number of flight lines, improving operational efficiency over expansive PV installations (e.g., those exceeding 100 MW of capacity).

Post-flight, all imagery was transferred to our cloud-based platform for ingestion, quality checks, orthomosaic construction, and georeferencing. More than 6,000 solar farms were collected across the U.S.; the geographic distribution of these sites can be seen in Figure~\ref{fig:locations}. Final GSD of the orthomosaics varied between 2.2 and 6.5 cm. Samples of sites ranging in plant capacities from 1-400 MW can be found in Figure~\ref{fig:sample} and distribution statistics of acquired orthomosaics can be found in Figure~\ref{fig:stats}.

\vspace{-2ex}\paragraph{Anomaly / Defect Types}\hspace{-2ex}
Our labeling process involves the localization and detection of panels with the categories as seen in Figure~\ref{fig:defect-types}. Categories can range from affecting only a single panel to an entire table of panels.

\vspace{-2ex}\paragraph{Data Labeling}\hspace{-2ex}
Data labeling was performed at the orthomosaic level in QGIS, with annotations saved in GeoJSON to preserve geospatial accuracy. This site-wide method, chosen over tile-based annotation, improved efficiency for large solar facilities. Defects were categorized using thermal signatures and predefined temperature thresholds, and automated quality checks—cross-referencing historical data—minimized errors. This structured approach establishes a solid foundation for training machine learning models and enables scalable, data-driven analysis of PV performance.

\vspace{-2ex}\paragraph{Data Diversity}\hspace{-2ex}
Data distribution is a critical factor in the dataset creation process. Sites are sourced from various locations across the North America region, requiring consideration of factors such as site locations, installation types (rooftop, ground-mounted), module types (poly-crystalline, thin film), module layouts (tracker, fixed), and prevailing weather and temperature conditions. This comprehensive approach was essential to ensure our pipeline's effectiveness across potentially unseen capture conditions.

\section{Architecture}

Our architecture, displayed in Figure~\ref{fig:architecture}, consists of several preprocessing steps and machine learning models which utilize each other's outputs to improve final defect detection performance and filter false positives.

\vspace{-2ex}\paragraph{Pre-Processing}\hspace{-2ex}
Orthos are processed with linear stretching normalization and outlier filtering using the ortho-level statistics. We then tile the images into $1024 \times 1024$ patches with 25\% overlap and perform an additional local tile-level linear stretching normalization and clipping to be in the range [0, 1]. To further enhance cross-region generalization performance, models are trained on a diverse dataset spanning multiple geographic regions, panel types, and mounting configurations.

\begin{table*}[t!]
\centering
\resizebox{0.85\textwidth}{!}{%
\begin{tabular}{@{}cccccc@{}}
\toprule
\textbf{Task}                    & \textbf{Architecture}                & \textbf{Backbone} & \textbf{\# Params (M)} & \textbf{Metric} & \textbf{Performance} \\ \midrule
\multirow{3}{*}{Table Detection}   & \multirow{3}{*}{U-Net~\cite{ronneberger2015u}}  & ResNet-18       & 14.3 & \multirow{3}{*}{mIoU}   & 96.1          \\
                                   &                         & ResNet-50       & 32.5 &    & 97.8          \\
                                   &                         & EfficientNet-B0 & 6.3 &    & \textbf{97.9} \\
\midrule
\multirow{3}{*}{Panel Detection} & \multirow{3}{*}{Faster R-CNN~\cite{ren2015faster}} & ResNet-50 DC5     & 166                    & \multirow{3}{*}{mAP@75}          & 78.3                 \\
                                   &                         & ResNet-50 FPN   & 42.0   &        & 79.8          \\
                                   &                         & ResNet-101 FPN  & 61.0   &        & \textbf{83.3} \\
\midrule
\multirow{3}{*}{Anomaly Detection} & \multirow{3}{*}{YOLOv5~\cite{redmon2016you}} & YOLOv5-S        & 7.2  & \multirow{3}{*}{Precision@50}  & 81.6          \\
                                   &                         & YOLOv5-M        & 21.2 &   & 85.0          \\
                                   &                         & YOLOv5-L        & 46.5 &   & \textbf{89.8} \\ 
\bottomrule
\end{tabular}%
}
\caption{\textbf{Empirical test set results of the individual model components in our aerial infrared solar farm inspection pipeline.} We report mean intersection-over-union (mIoU) for Table Detection which is posed as a binary segmentation task. We report mean Average Precision (mAP) for Panel Detection which is posed as a oriented object detection task. We report Precision for Anomaly Detection which is posed as a horizontal object detection task.}
\label{tab:model-results}
\end{table*}
\vspace{-2ex}\paragraph{Table Detection}\hspace{-2ex}
We first detect solar farm tables, a grouping of connected panels, utilizing a U-Net architecture~\cite{ronneberger2015u} with a ResNet-50 backbone~\cite{he2016deep} to perform binary segmentation of table vs. background. We post-process predictions to rectangular polygons, which are then utilized downstream to filter any off-table false positive predictions. We further use these table localizations to filter out patches without panels to reduce the number of processed patches in our pipeline, as well as renormalize the images by filtering high temperature outliers in the background.

\vspace{-2ex}\paragraph{Panel Detection}\hspace{-2ex}
The panel detection model employs an oriented object detection approach to accurately localize individual panels. Unlike segmentation-based methods, which struggle to differentiate closely spaced panels, we utilize a rotated Faster R-CNN architecture~\cite{xie2021oriented, ren2015faster} with a ResNet-50 Feature Pyramid Network (FPN) backbone~\cite{he2016deep,lin2017feature}. This approach enables precise detection of panel corners with varying orientations, mitigating errors caused by perspective distortions and alignment inconsistencies in the captured orthomosaics.

We convert the detected oriented boxes to polygons and post-process erroneous rotation predictions by averaging the rotation angle of all panels in a table and then snapping the angles to the orientation of the corresponding table.

\vspace{-2ex}\paragraph{Anomaly Detection}\hspace{-2ex}
The anomaly detection model is a key component of our pipeline, responsible for identifying defective PV cells exhibiting abnormal thermal signatures. We employ a YOLOv5-based object detection architecture~\cite{redmon2016you}, which has been optimized to detect various types of panel defects, including hot spots, string outages, and diode failures. Given the variability in temperature distributions across different solar farms, our model incorporates temperature normalization techniques such as local histogram equalization and adaptive thresholding to improve robustness. The model is trained on a geographically diverse dataset, ensuring generalization across different climates and panel types. Achieving a peak Precision@50 of 89.8\%, our system effectively localizes anomalies while minimizing false positives through post-processing steps that filter detections occurring outside predefined panel regions. These high-confidence detections enable rapid assessment of PV farm health, reducing the need for extensive manual inspections and facilitating proactive maintenance interventions.

\begin{figure}[ht!]%
    \centering
    \begin{subfigure}[b]{0.325\linewidth}
        \includegraphics[width=\linewidth]{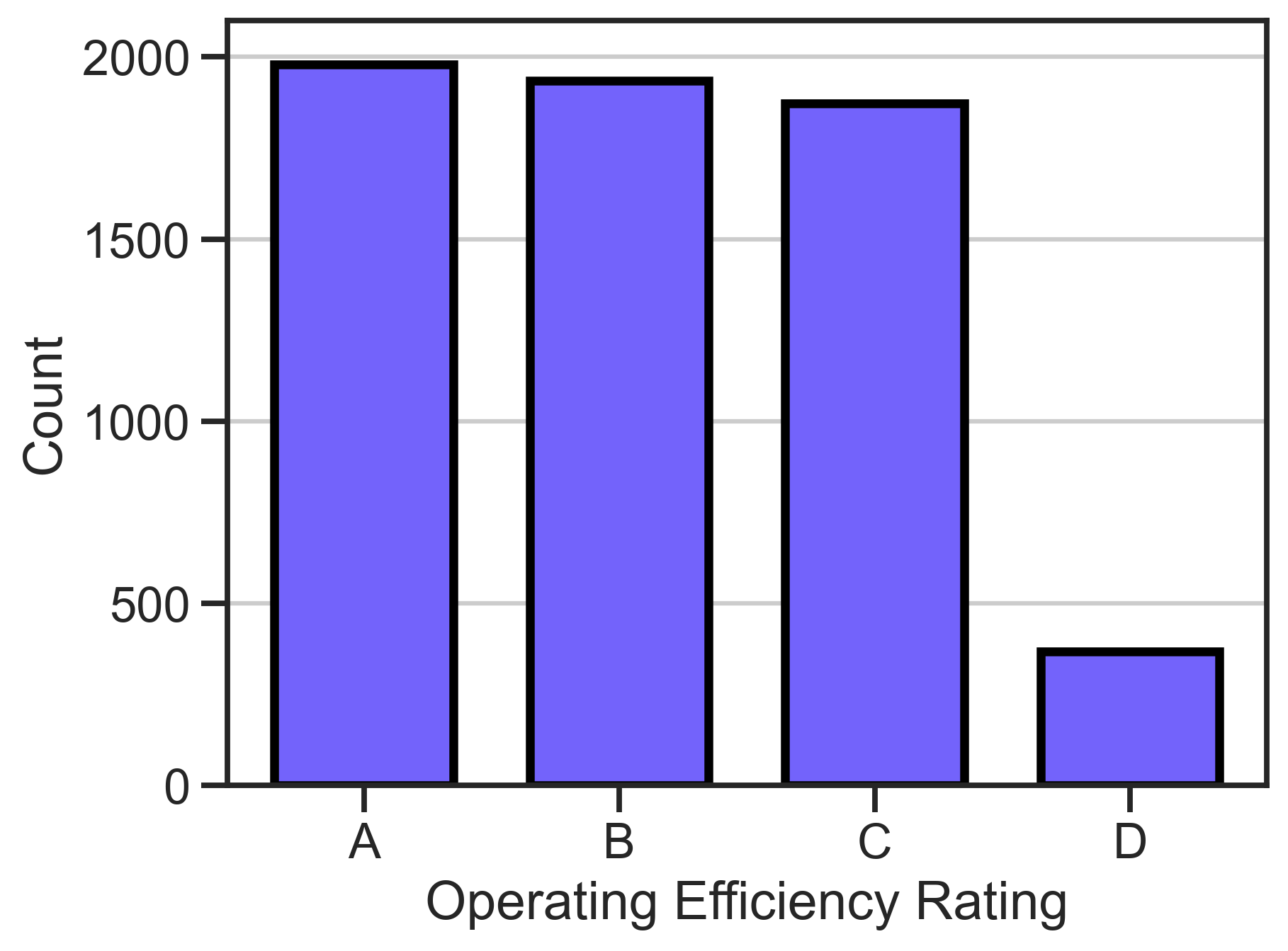}
        \caption{}
    \end{subfigure}
    \begin{subfigure}[b]{0.325\linewidth}
        \includegraphics[width=\linewidth]{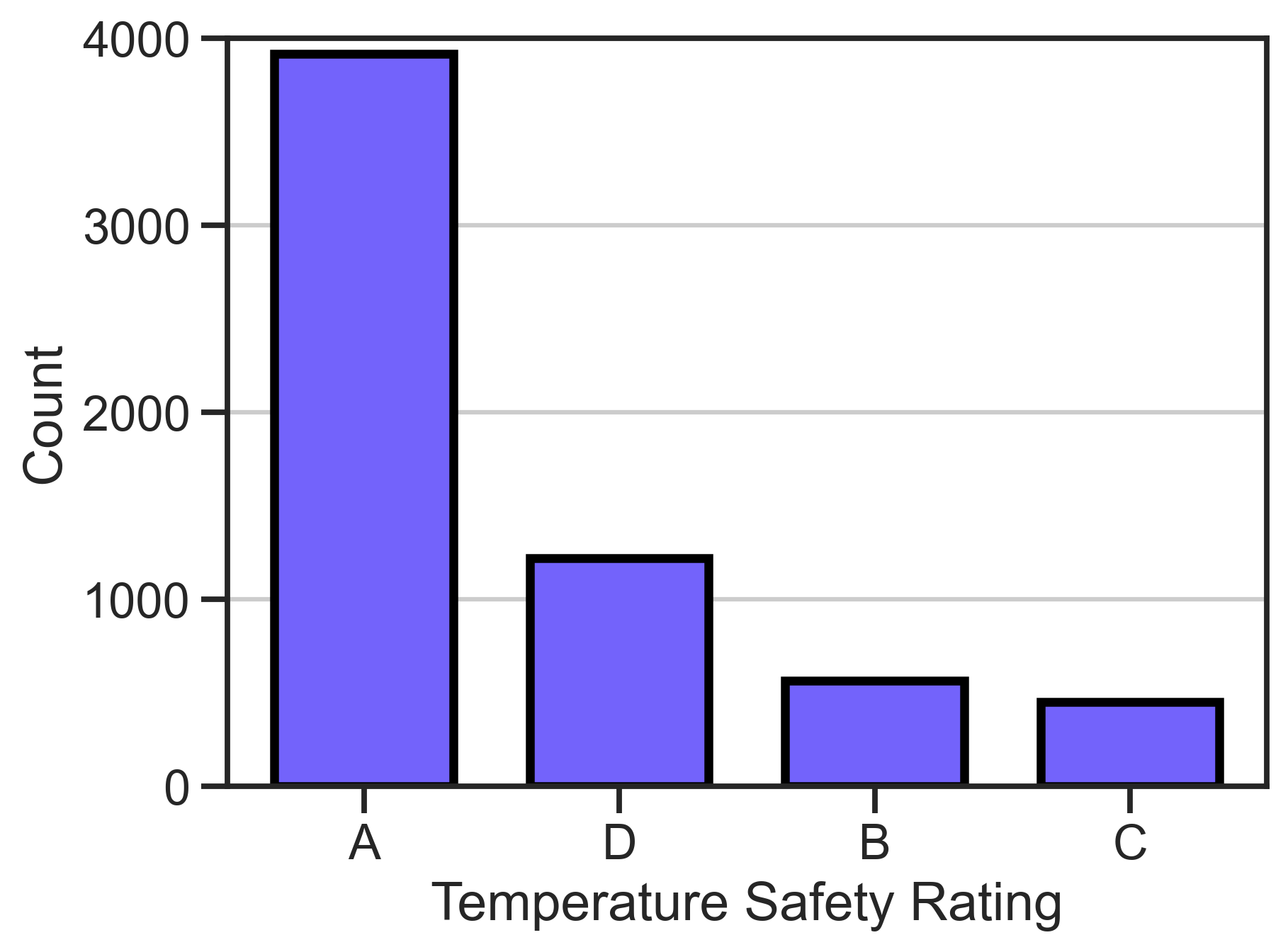}
        \caption{}
    \end{subfigure}
    \begin{subfigure}[b]{0.325\linewidth}
        \includegraphics[width=\linewidth]{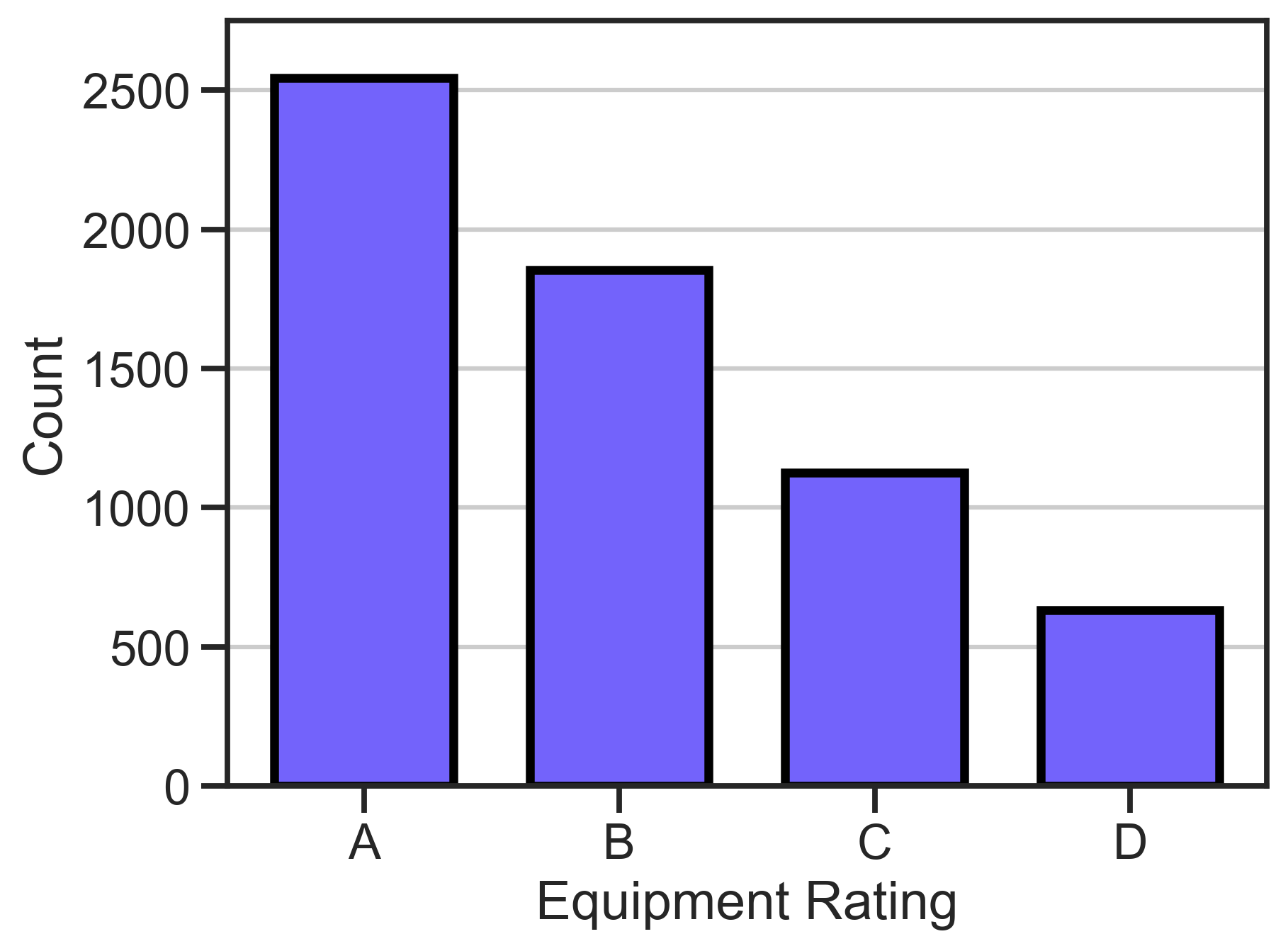}
        \caption{}
    \end{subfigure}
    \\
    \begin{subfigure}[b]{0.325\linewidth}
        \includegraphics[width=\linewidth]{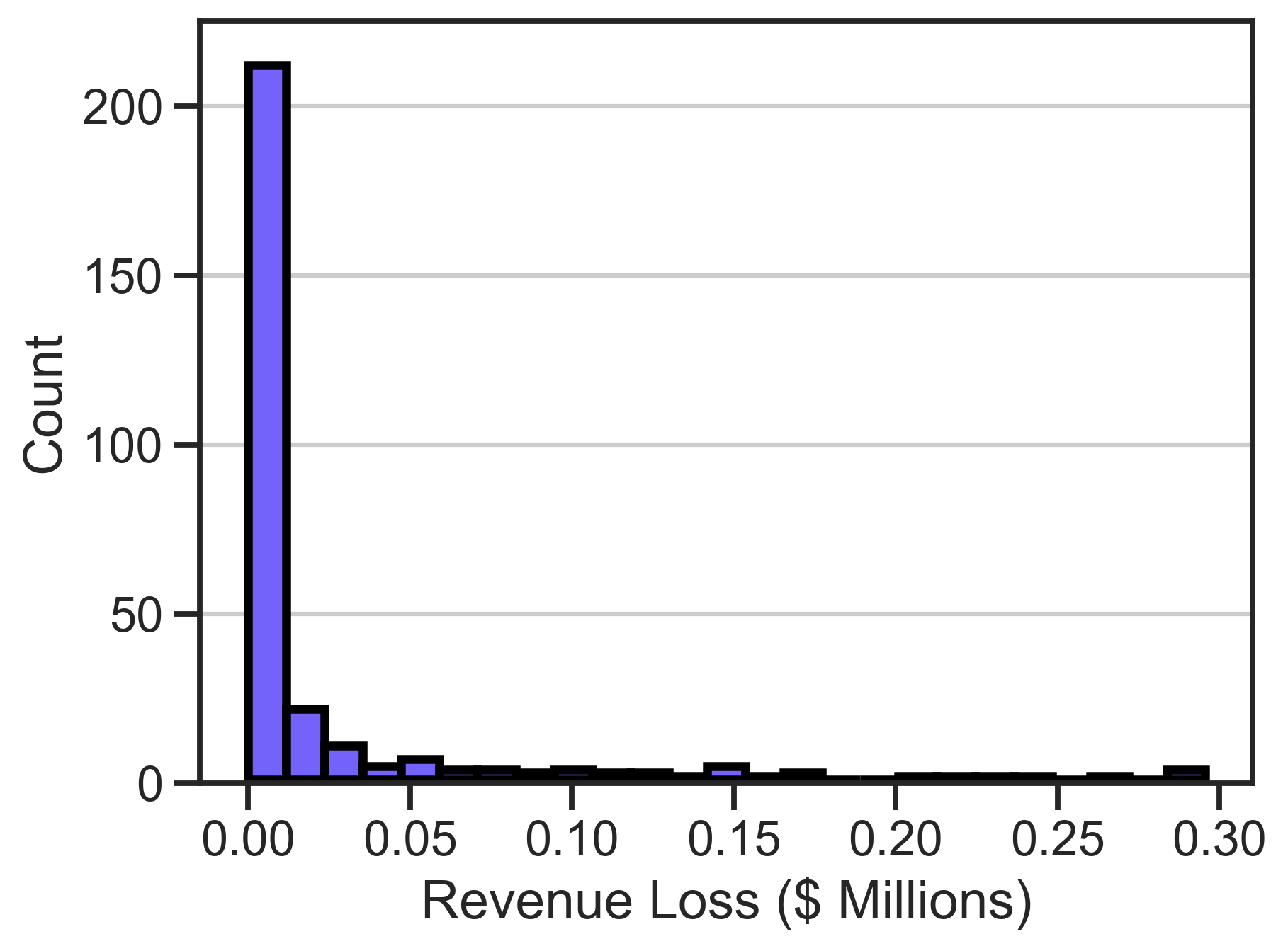}
        \caption{}
    \end{subfigure}
    \begin{subfigure}[b]{0.325\linewidth}
        \includegraphics[width=\linewidth]{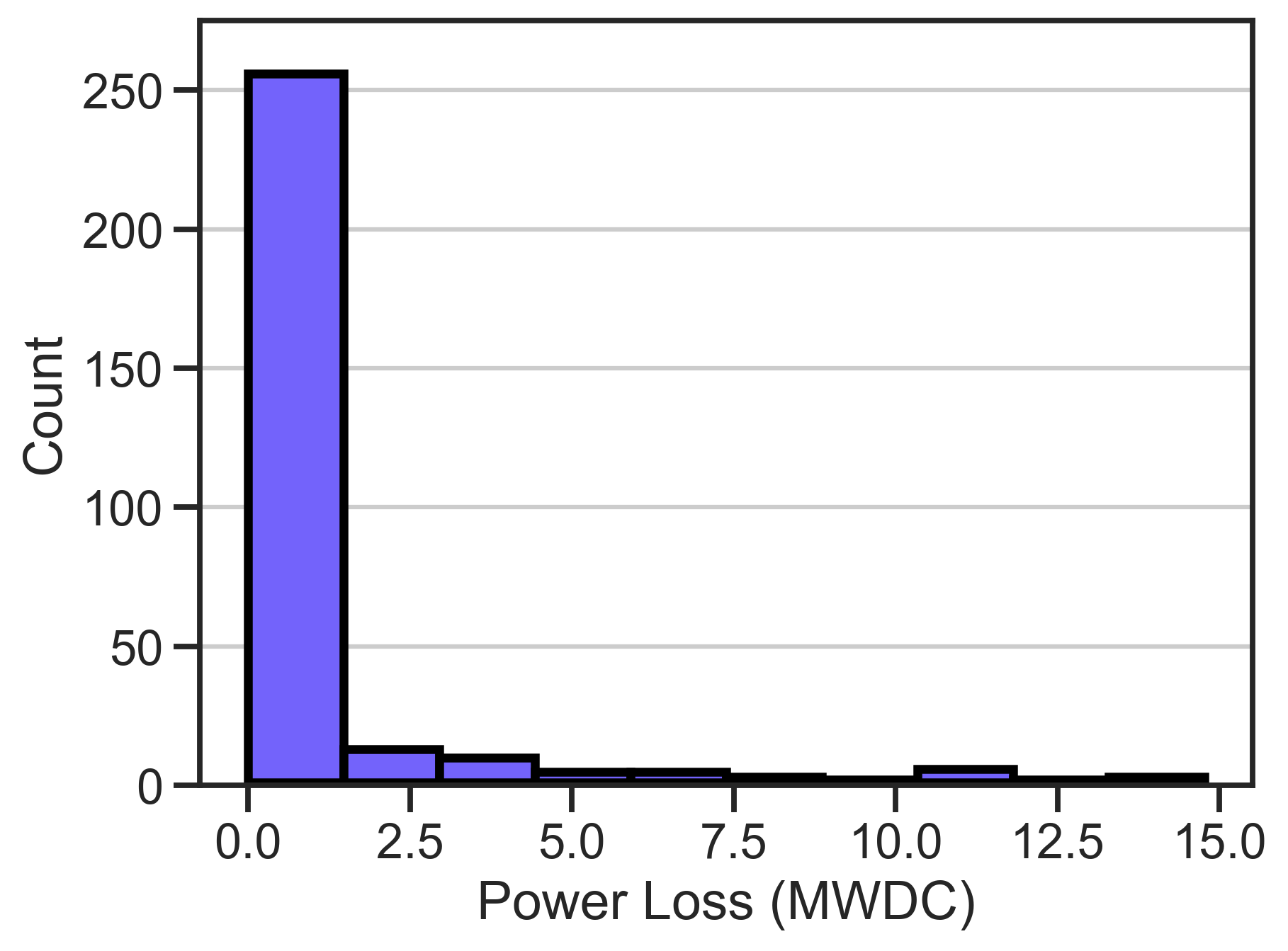}
        \caption{}
    \end{subfigure}
    \begin{subfigure}[b]{0.325\linewidth}
        \includegraphics[width=\linewidth]{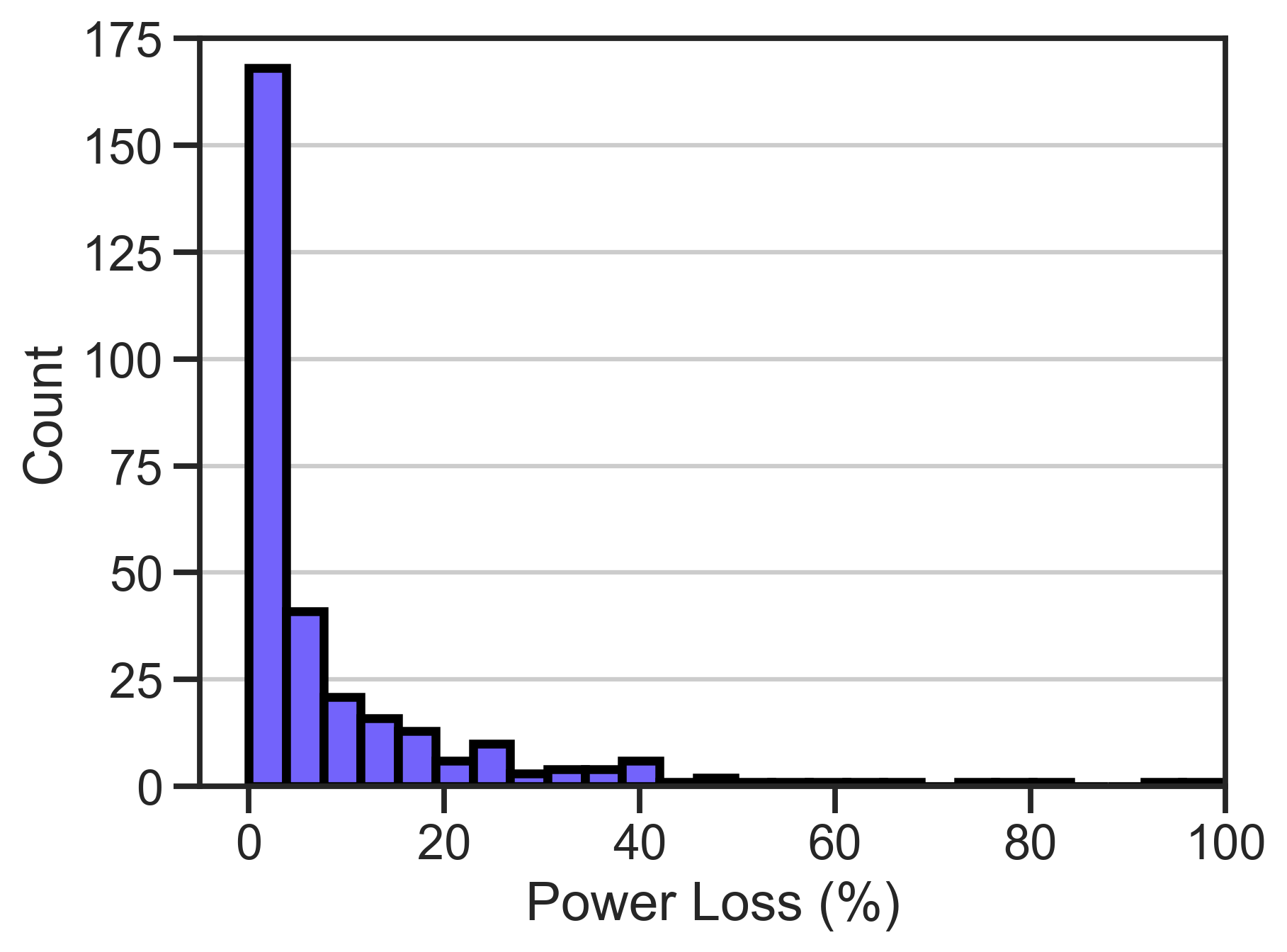}
        \caption{}
    \end{subfigure}
    \caption{\textbf{Analytics Results of our Large-Scale Airborne Infrared Scan.} Distribution plots of (a) operational efficiency rating, (d) estimated revenue loss in millions (\$) (e) total power loss per site in MWDC, and (f) \% power loss per site.}
    \label{fig:analysis}%
\end{figure}
\begin{figure*}[t!]
\centering
\includegraphics[width=0.95\linewidth]{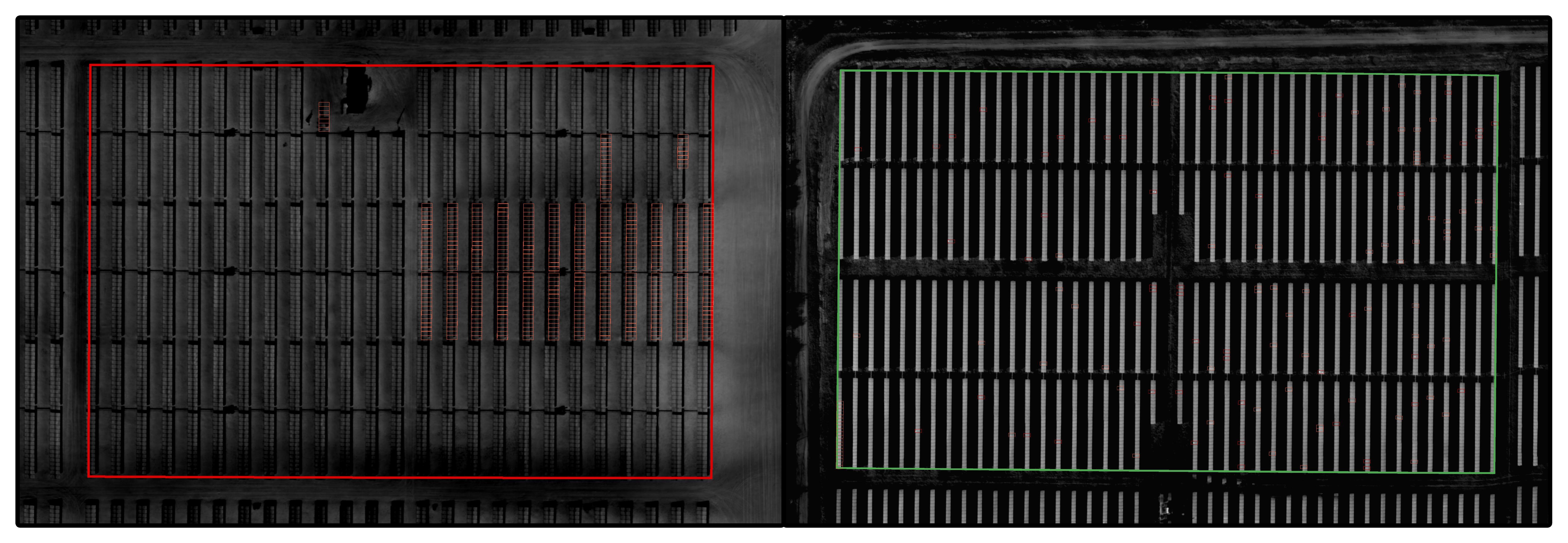}
\caption{\textbf{Sample anomaly detection outputs of our Infrared Solar PV Plant Inspection Pipeline.} Using custom QGIS plugins our analysts see inverter blocks separated with green/red boxes. Each defective panel is identified with a box around the anomalous area. Later in our pipeline, anomaly predictions are merged with panel detections to produce tighter boxes around each panel. \textit{Best viewed while zoomed in.}}
\label{fig:samples-ui}
\end{figure*}

\vspace{-2ex}\paragraph{Safety Hazard Detection}\hspace{-2ex}
The safety hazard detection model identifies high-temperature anomalies within panels, where excessive heat can indicate efficiency losses or fire risks. A hotspot is defined as a localized temperature anomaly exceeding a 5$\degree$C threshold relative to neighboring cells. To accurately capture these defects while minimizing noise, we employ a max-median thresholding approach, subdividing each panel into smaller grids. A region is classified as a hotspot if the difference between its maximum and median temperature exceeds a predefined threshold, ensuring robustness across varied environmental conditions.

To mitigate false positives near panel edges—often caused by segmentation inaccuracies or elevated background temperatures—we implement context-aware boundary padding, filtering out detections in a buffer zone while balancing the risk of false negatives. Hotspots are categorized into five severity levels based on temperature differentials, $\Delta T$, ranging from low-risk, $\Delta T < 5\degree C$, to critical-risk, $\Delta T \geq 15\degree C$. This structured classification enables targeted maintenance prioritization, helping operators proactively address potential failures before they escalate.

\vspace{-2ex}\paragraph{Post-Processing}\hspace{-2ex}
Post-processing integrates outputs from the table detection and panel detection models to refine anomaly predictions and perform statistical temperature analysis of panels. First, detected anomalies are filtered by cross-referencing them with panel locations, ensuring that only defects within valid panel regions are retained while discarding false positives from background areas such as roads or vegetation. Additionally, temperature statistics are computed for each detected panel, including median, mean, and maximum temperature deviations from the site baseline. These statistics help differentiate between natural environmental variations and true defects, improving the reliability of automated assessments. The final processed data is structured in GeoJSON format, enabling seamless integration with downstream inspection and maintenance workflows. This approach enhances defect localization accuracy while providing actionable insights into PV system performance at scale.

\begin{table*}[ht!]
\centering
\resizebox{1.0\textwidth}{!}{%
\begin{tabular}{@{}cccc@{}}
\toprule
\textbf{Rating} & \textbf{Operating}          & \textbf{Temperature Safety} & \textbf{Equipment}        \\ \midrule \addlinespace
\textbf{$A$}      & $OR \geq 99.5\%$               & $\Delta T_{Max} < 10\degree C$        & $0 \leq APM < 12.9$    \\\addlinespace
\textbf{$B$}      & $97.5\% \leq OR < 99.5$                & $10\degree C \leq \Delta T_{Max} < 15\degree C$ & $13 \leq APM < 51.9$   \\\addlinespace
\textbf{$C$}      & $80\% \leq OR < 97.5\%$ & $15\degree C \leq \Delta T_{Max} < 20\degree C$ & $52 \leq APM < 172.9$ \\\addlinespace
\textbf{$D$}      & $OR \leq 79.9\%$                & $\Delta T_{Max} \geq 20\degree C$                & $APM > 173$      \\ \addlinespace\midrule\midrule \addlinespace
\textbf{Definitions:} &
  \begin{tabular}[c]{@{}c@{}}$C_{Total}: \textit{Total Plant Capacity (MW)}$\\\addlinespace $C_{Defect}: \textit{Total Capacity of Defective Modules (MW)}$\\\addlinespace $OR \textit{ (Operational Ratio)} = (C_{Total} - C_{Defect}) / C_{Total}$\end{tabular} &
  \begin{tabular}[c]{@{}c@{}}$T_{Hotspot}: \textit{Temp. of a given Hotspot (MW)}$\\\addlinespace $T_{Normal}: \textit{Normal Temp. of Non-Defective Modules (MW)}$\\\addlinespace $\Delta T = T_{Hotspot} - T_{Normal}$\end{tabular} &
  \begin{tabular}[c]{@{}c@{}}$C_{Total}: \textit{Total Plant Capacity (MW)}$\\\addlinespace$A_{Total}: \textit{Total \# of Defective Modules}$\\\addlinespace $APM = A_{Total} / C_{Total}$\end{tabular} \\ \addlinespace \bottomrule
\end{tabular}%
}
\caption{\textbf{Definitions of our Solar PV Plant Health Ratings.} Our ratings consist of a combination of \textbf{Operating Efficiency Rating} (estimated DC loss due to detected hot cells), \textbf{Temperature Safety Rating} (delta temperature between hot and normal functioning cells), and \textbf{Equipment Rating} (hot cells per MW in comparison to solar plants of similar age and module type).}
\label{tab:algorithm}
\end{table*}

\vspace{-2ex}\paragraph{Deployment Details}\hspace{-2ex}
Our processing pipeline is optimized for scalability and efficiency, leveraging AWS Batch with cost-effective CPU instances and Metaflow to orchestrate DAG-based AWS Step Functions. Orthomosaic patches are read directly from Cloud Optimized GeoTIFFs (COG) stored in AWS S3, minimizing I/O overhead and enabling on-the-fly processing without full file downloads. While processing a 10 km\textsuperscript{2} solar site orthomosaic ($\sim$50GB) takes approximately 3 hours, and a 1 km\textsuperscript{2} site ($\sim$5GB) completes in 20 minutes, this compute time is nearly negligible compared to the total latency of data capture and upload, which can take several hours to days depending on site connectivity.

\begin{figure*}[t!]
\centering
\includegraphics[width=0.9\linewidth]{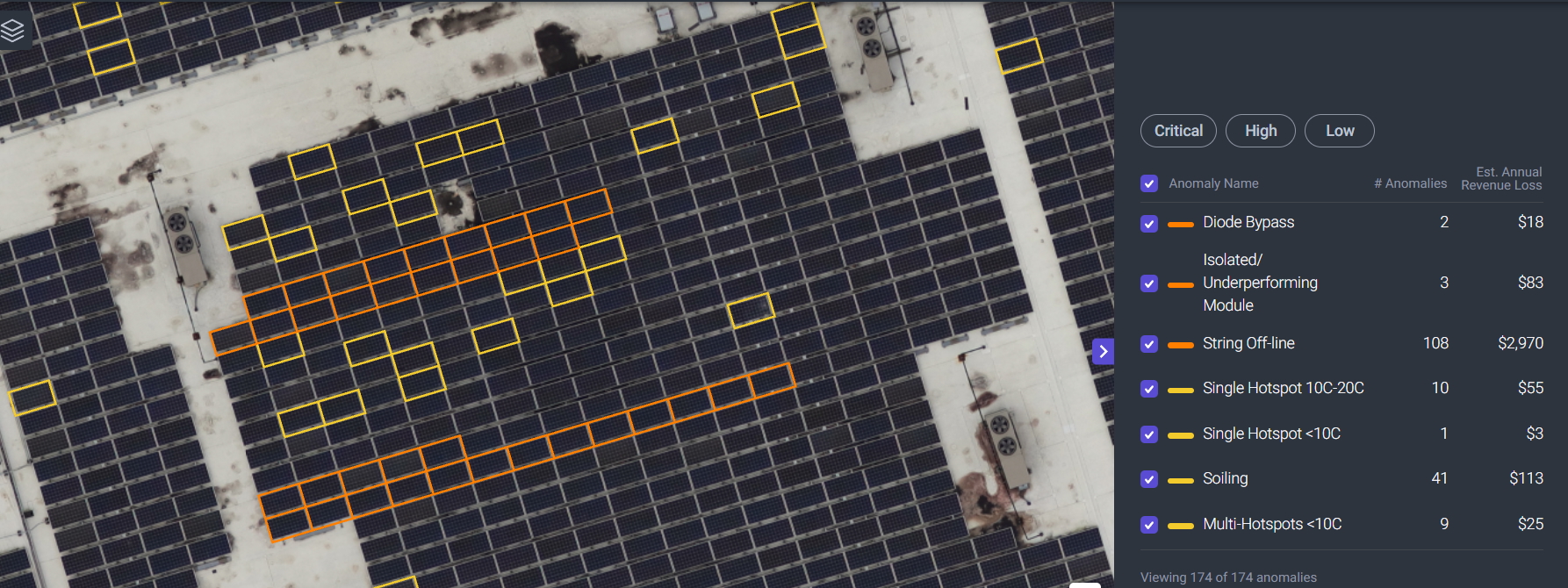}
\caption{\textbf{Solar PV Plant Health Monitoring User Interface (UI).} Our UI allows for toggling between aligned RGB and Infrared orthomosaics of the plant. Each panel with a detected defect is highlighted with \textcolor{red}{\textbf{red}}/\textcolor{orange}{\textbf{orange}}/\textcolor{Goldenrod}{\textbf{yellow}} box representing \textcolor{red}{\textbf{critical}}/\textcolor{orange}{\textbf{high}}/\textcolor{Goldenrod}{\textbf{low}} severity, respectively. Each defective panel has an estimated power and revenue loss which is summarized in the right pane.}
\label{fig:samples-ui-zoom}
\end{figure*}

Since our pipeline processes data asynchronously as it becomes available, the primary bottleneck remains aerial data acquisition rather than computational throughput, ensuring rapid turnaround once imagery is accessible. Some sample outputs of our final pipeline are provided in Figure~\ref{fig:samples-ui}. We further develop a UI which visualizes infrared and RGB orthomosaics such that defects can be correlated across modalities. A sample of detected infrared defects overlaid onto an RGB orthomosaic is visualized in Figure~\ref{fig:samples-ui-zoom}.

\section{Results}
Our individual model results are provided in Table~\ref{tab:model-results}. Our Table Detection model tends to generalize well across diverse variations in solar farms, achieving a mIoU of \textit{97.9\%} on the geographically split test set. We find that this model can accurately digitize entire sites with minimal manual corrections. These results improve the full pipeline's ability to filter false positive detections from downstream models.

The Panel Detection model achieves an impressive \textit{83.3\%} mAP@75 using a Rotated Faster R-CNN architecture. While it is common to report mAP@50 IoU, we find that a higher 75\% IoU is highly correlated with a positive qualitative assessment from our analyst team.

The Anomaly Detection model, leveraging the YOLOv5 architecture, achieved a high precision score of up to \textit{89.8\%}, demonstrating its robustness in identifying hotspots and faulty panels. However, challenges remain in differentiating between environmental factors (such as soil temperature and shading effects) and true panel defects. The integration of temperature normalization techniques helped mitigate these issues, but further refinement may improve performance in varied geographic and seasonal conditions.

Anecdotally, we find that Vision Transformer-based models~\cite{dosovitskiy2020image} do not yield significant performance improvements over the models in Table~\ref{tab:model-results}, while also increasing training complexity and computational cost.

\section{Analysis}

\subsection{Health Rating System}
Through thermal imaging and anomaly detection, each asset is assigned a three-letter rating (Table~\ref{tab:algorithm}) reflecting key performance attributes and derived from an adaptation of the standard bond scale (e.g., \texttt{AAA} to \texttt{DDD}). Letter grades (\texttt{A}, \texttt{B}, \texttt{C}, \texttt{D}) indicate asset condition based on industry best-known methods and practices, enabling rapid triage for deeper analysis. This framework aids stakeholders in comparing asset health across regions, technologies, and service providers, facilitating strategic decisions about investment, maintenance, and vendor relationships. The following are definitions of our three-rating system:

\vspace{-2ex}\paragraph{Operator Rating}\hspace{-2ex}
Estimated percentage DC losses.  This was derived by estimating standardized power losses attributed to known IR signatures for each polycrystalline or thin-film solar module and aggregating losses against known production figures or estimated production figures based on industry metadata. 

\vspace{-2ex}\paragraph{Temperature Rating}\hspace{-2ex}
Highest temperature delta within a given module and asset characteristics.  Utility-scale or large-scale PV plants were evaluated on a different scale than smaller rooftop sites where the hot cell temperature differential would have a higher impact on the potential safety risk of a fire.  

\vspace{-2ex}\paragraph{Equipment Rating}\hspace{-2ex}
Anomalies per megawatt peak (DC).  The total number of anomalies detected was normalized based on the PV plant's size to understand the number of module-specific anomalies.  String outages were removed from this calculation. 

\begin{figure}[t!]
\centering
\includegraphics[width=0.9\linewidth]{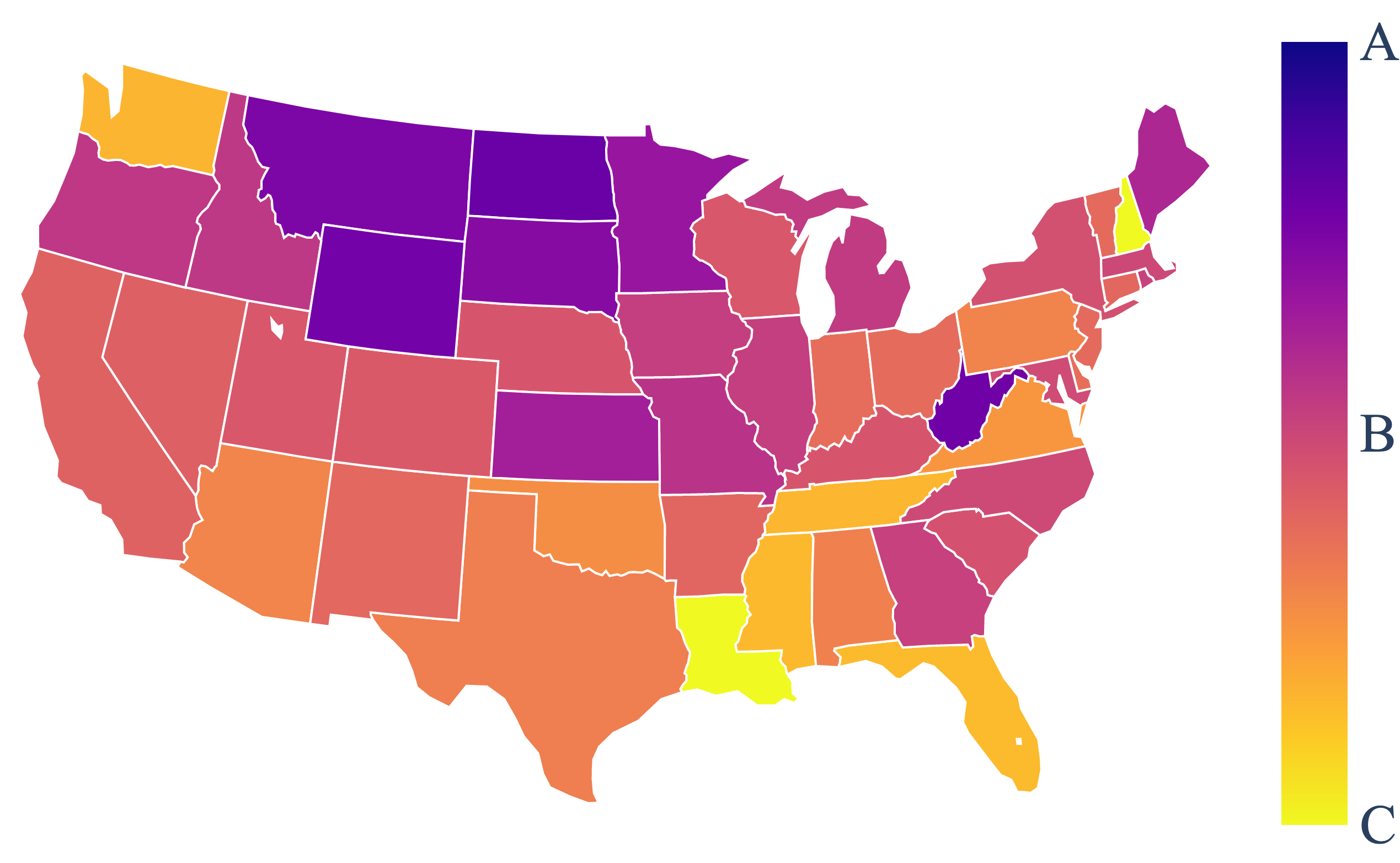}
\caption{\textbf{Mean Operational Efficiency Rating by State}. We average the operational efficiency ratio by state and plot as a heatmap where darker colors indicate \textcolor{Violet}{\textbf{higher effiency}} and brighter indicates \textcolor{YellowOrange}{\textbf{lower effiency}}.}
\label{fig:ratings-by-state}
\end{figure}

\subsection{North American Scan Analysis}

Across our large-scale aerial infrared inspections of 6,155 solar farms in the United States, we observe a multifaceted performance landscape shaped by site age, mounting configurations, and plant size. Notably, 51\% of the surveyed U.S. fleet currently maintains excellent or good operational conditions, corresponding to asset ratings of BBB or higher, demonstrating robust performance across a wide geographic spread. Below we discuss several patterns evident in our analyses.

\vspace{-2ex}\paragraph{Aging Dynamics}\hspace{-2ex} Despite the natural performance decay expected from plants exceeding ten years of service, approximately 25\% of these older sites continue to exhibit excellent or good operational health, with 11\% of that subset holding an AAA rating. This AAA status implies minimal degradation in each of the three assessment areas—thermal, electrical, and mechanical—and highlights the efficacy of proactive maintenance strategies in prolonging system life and retaining superior performance.

\vspace{-2ex}\paragraph{Influence of Plant Size and Mounting}\hspace{-2ex} Larger installations ($\geq$100 MW) demonstrate a particularly strong operational profile, with over 50\% of such sites rated as excellent or good. Moreover, tracker-mounted systems fare better over time than fixed-tilt sites; 54\% of trackers sustain a good or better rating in contrast to 44\% of fixed-tilt counterparts. These observations suggest that larger plants and tracking technologies may be more resilient to environmental stressors and mechanical wear, underscoring the value of informed technology choices for long-term asset performance.

\vspace{-2ex}\paragraph{Geographic Variability and Industry-Wide Losses}\hspace{-2ex} As illustrated in Figure~\ref{fig:ratings-by-state}, certain states with older infrastructure or more extreme environmental conditions continue to register higher defect rates and larger energy losses, correlating with lower operational efficiency ratings. Southern regions, in particular, show a notable concentration of C-level ratings, although exceptions exist. Conversely, states with a higher prevalence of newer builds or concentrations of sites under management with more rigorous maintenance regimens enjoy significantly lower defect densities.

\section{Limitations, Challenges, \& Future Work}

In comparison to the visible spectrum, infrared scans experience significant variance in the recorded temperature values based on several factors such as acquisition time of day, seasonality, and weather. Furthermore, unique soil types and their temperatures vary across the U.S. When these temperatures are similar to the temperatures of solar panels or anomalies, distinguishing the defects of interest can become difficult to distinguish from the background.

These variables make selecting global optimal statistics for normalization of temperatures infeasible. Due to this, we utilize a pipeline of \textit{ortho-level outlier clipping}, \textit{linear stretching}, \textit{tiling with overlap}, \textit{tile-level min-max normalization}, and \textit{histogram equalization} to preprocess the input images to our machine learning models. 

Furthermore, while it is debatable whether to use a single anomaly detection model for solar plant health monitoring instead of our cascaded model, we find that our approach provides a significant improvement in false positive and false negative reductions. Additionally, utilizing common architectures for each problem type allows for more seamless training and deployment updates to our pipeline.

For future work, we plan to investigate the feasibility of creating an image-level pipeline instead of ortho-level, which can be deployed at the edge on a sUAS or drone. This would allow for any immediate safety risks to be identified quickly before the pilot even finishes data acquisition of the site of interest. Additionally, we plan to investigate domain adaptation techniques~\cite{tuia2021recent} to streamline normalization and reduce reliance on site-specific statistics.

\section{Conclusion}
\label{sec:conclusion}
This work demonstrates the effectiveness of large-scale aerial infrared inspections combined with machine learning for monitoring the health of PV farms. By leveraging a geographically diverse dataset and an automated deep-learning pipeline, we enable precise defect detection, enhancing operational efficiency and optimizing maintenance strategies. Our health rating system provides a standardized assessment, allowing asset owners to make informed decisions and mitigate revenue losses due to undetected anomalies.

Despite advancements, challenges remain in temperature normalization and reducing false positives from environmental factors. Future improvements will focus on integrating time-series performance data, enhancing model generalization across climates, and refining post-processing techniques. As solar energy continues to expand, AI-driven monitoring solutions like ours will be crucial in ensuring the reliability and efficiency of renewable energy infrastructure.

{
    \small
    \bibliographystyle{ieeenat_fullname}
    \bibliography{main}
}

\end{document}